\documentclass[submission,copyright,creativecommons]{eptcs}
\providecommand{\event}{FMAS 2020} 
\usepackage{breakurl}             
\usepackage{underscore}  
\usepackage[table]{xcolor}
\usepackage{subcaption}
\usepackage[labelformat=parens,labelsep=quad,skip=3pt]{caption}
\usepackage{cite,xcolor,textcomp,graphicx,tabularx, algorithmic,amsmath,amssymb,amsfonts,hyperref,todonotes,dsfont,tabularx,booktabs,hyperref,cleveref,tikz,marvosym,multirow,textcomp,commands,enumitem,tabularx}
\title{How to Formally Model Human in Collaborative Robotics}
\author{Mehrnoosh Askarpour
\institute{McMaster University\\ Canada}
\email{askarpom@mcmaster.ca}
}

\def\titlerunning{How to Formally Model Human in Collaborative Robotics}
\def\authorrunning{M. Askarpour}

\begin{document}
\maketitle

\begin{abstract}
Human-robot collaboration (HRC) is an emerging trend of robotics that promotes the co-presence and cooperation of humans and robots in common workspaces.  
Physical vicinity and interaction between humans and robots, combined with the uncertainty of human behaviour, could lead to undesired situations where humans are injured. Thus, safety is a priority for HRC applications.

Safety analysis via formal modelling and verification techniques could considerably avoid dangerous consequences, but only if the models of HRC systems are comprehensive and realistic, which requires reasonably realistic models of human behaviour. This paper explores state-of-the-art solutions for modelling human and discusses which ones are suitable for HRC scenarios.
\end{abstract}

\section{Introduction}
A new uprising section of robotics is Human-Robot Collaboration (HRC), where human operators are not dealing with robots only through interfaces, but they are physically present in the vicinity of robots, performing hybrid tasks (i.e., partially done by the human and partially by the robot).
These applications introduce promising improvements in the industrial manufacturing area by combining human flexibility and machine productivity~\cite{Matthias2015}, but they must assure the safety of human operators before being fully deployed to certify that interactions with robots will not cause any harm or injury to humans.

Formal methods have been widely used in robotics for decades in a variety of applications including mission planning~\cite{meng,askarpourmind,R201869,DBLP:conf/aips/CrosbyPRK17,9121678}, 
formal verification of properties~\cite{DBLP:journals/sosym/MiyazawaRLCTW19,DBLP:journals/csur/LuckcuckFDDF19,DBLP:conf/ifm/LuckcuckFDD019,FM2016}, and controllers~\cite{xyz,DESILVA2019130,DBLP:journals/fac/BersaniSMPR20}. They could be an effective means for the safety analysis of traditional and collaborative robotics due to their comprehension and exhaustiveness~\cite{GUIOCHET201743,DBLP:journals/trob/VicentiniARM20,VISINSKY1994139}. 
However, a formal model of an HRC system should also reflect the human factors that impact the state of the model without dealing with the additional details of human mental and emotional processes. 
It necessitates building a formal model of human behaviour that replicates their physical presence and the observable manifestation of their behaviour which includes both executing the required job following the expected instructions, and deviations from the expected behaviour (i.e., mistakes, errors, malicious use). Thus, the model of the human for each HRC scenario might be intertwined with the model of the executing job. \\
A 100\% realistic formal model of human might not be a reasonable goal and, just like any other phenomena, any human model is subject to some level of abstraction and simplification. 
Moreover, unlike robots that perform only a limited set of activities, all possible activities of the human are not foreseeable. 
Besides, humans are non-deterministic, and a realistic algorithm for their behaviour is not easy to envision. 

To tackle these issues, researchers have explored well-established cognitive investigations, task-analytic models, and probabilistic approaches~\cite{BoltonBS13}. These three tracks are not mutually exclusive; for example, there are instances of cognitive probabilistic models in the literature that will be discussed later in the paper.
The rest of this paper, reports on the state-of-the-art on each of these possibilities and examines their compatibility for HRC scenarios, following a snowballing literature review.
Moreover, for each track there are several instances that adhered to the normative human behaviour, while other instances considered erroneous behaviour too; hence, a separate section is dedicated to examples that also model errors. 

In addition to the three highlighted possibilities, Bolton et al.~\cite{Bolton:2012} considered human-device interface models, which are out of the scope of this paper. They do not consider the physical co-presence of humans and robots and focus only on their remote communication; thus, human physical safety has never been an issue in these studies \cite{Bolton2010}.

The rest of this paper is structured as follows: \Cref{sec:cog} explores the cognitive approaches; \Cref{sec:ta} reviews task analytic approaches; \Cref{sec:prob} discusses the probabilistic techniques; \Cref{sec:err} specifically discusses the difficulties of modelling human errors; finally \Cref{sec:con} draws a few conclusions on the best-suited solution for HRC scenarios.

\section{Cognitive Models}
\label{sec:cog}
Cognitive models specify the rationale and knowledge behind human behaviour while working on a set of pre-defined tasks. They are often incorporated in the system models and contain a set of variables that describe the human cognitive state, whose values depend on the state of task execution and the operation environment~\cite{BoltonBS13}. Famous examples of well-established cognitive models follow.

SOAR\cite{laird2012soar} is an extensive cognitive architecture, relying on artificial intelligence principles, that reproduce human reasoning and short and long term memories. However, it only permits one operation at a time which seems not to be realistic in HRC scenarios (e.g., human sends a signal while moving towards the robot).

ACT-R \cite{anderson1996act} is a detailed and modular architecture that depicts the learning and perception processes of humans. It contains modules that simulate declarative (\ie known facts like $2 + 2 = 4$) and procedural (\ie knowledge of how to sum two numbers) aspects of human memory. An internal pattern matcher searches for the procedural statement relevant to the task that the human needs to perform (i.e., an entry in declarative memory) at any given time. SAL~\cite{lebiere2008sal,JilkLOA08} is an extended version of ACT-R, enriched with a neural architecture called Leabra. 

SOAR and ACT-R highlight the cognition behind erroneous behaviours of the human~\cite{CurzonB02} that could impact safety, such as over-trusting or a lack of trust in the system. It could be used to generate realistic models that also reflect a human deviation from the correct instructions.
On the other hand, they both lack a formal definition and cannot be directly inputted to automated verification tools.
Hence, they must be transformed into a formal model, which is a cumbersome and time-consuming process and requires extensive training for modellers ~\cite{Salvucci:2003}. 
Moreover, they are detailed and heavy models and, therefore, must be abstracted before formalization to avoid a state-space explosion.
There are examples of formal transformation of ACT-R in \cite{GallF14,both2007formal,DBLP:conf/cogsci/LangenfeldWP19} and of SOAR in \cite{HowesY97}. However, they remain dependent on their case-studies or use arbitrary simplifications, and therefore, cannot be re-used as a general approach. Thus, providing a trade-off between abstraction and generality of these two cognitive models is not an easy task.

Programmable User Models (PUM) define a set of goals and actions for humans. The model mirrors both human mental actions (i.e., deciding to pick an object) and physical actions (i.e., pick an object). These models have a notion of a human \textit{mental model} \cite{COGS334,RITTER20011} and separate the machine model from the user's perception of it \cite{Bredereke:2002}, that avoids mode confusions which happen when the observed system behaviour is not the same as the user's expectation \cite{Bredereke2005229}. Additionally, Moher et al.~\cite{Moher1995} assign a \textit{certainty} level to mental models whose different adjustments reveal various human reactions in execution situations. Curzon et al.~\cite{Curzon2007} introduce two customized versions of mental models as naive or experienced.

Since PUM has been around for so long, simplified formal versions of it are defined for a variety of applications such as domestic service robots~\cite{Stocker2012} Formalised with \cite{Brahm}, interactive shared applications~\cite{Butterworth2000} Formalised with logic formulae, and Air Management System~\cite{WertherS05} Formalised with Petri-nets.
PUM models have a general and accurate semantics and could be well-suited inputs for automated formal verification tools upon simplification. 
On the other hand, they are very detailed and large and risk the state-space exploration phenomena. As mentioned above, there are simplified versions of PUM models which again are not generic enough to be used for different scenarios including HRC applications. Hence, the effort and time required for customization and simplification remain as high as for SOAR and ACT-R.
\section{Task-Analytic Models}
\label{sec:ta}
Task-analytic models, as their name suggests, analyze human behaviour throughout the execution of the task. Therefore, they study the task 
as a hierarchy of atomic actions.
By definition, these models reflect the expected behaviour of the human, which leads to correct execution of the task, and do not focus on reflecting erroneous behaviour ~\cite{Bolton:2012}. Recently, however, Bolton et al.~\cite{BOLTON2019168} put together a task-based taxonomy of erroneous human behaviour that allows errors to be modelled as divergences from task models; Li et al.~\cite{LI2016318} uses an analytic hierarchy process to identify hazards and increase the efficiency of the executing task.

Examples of task-analytic models follow. Paterno et al.~\cite{paterno1997concurtasktrees} extend ConcurTaskTrees (CTT) \cite{ctt} to better express the collaboration between multiple human operators in air traffic control; Mitchell and Miller~\cite{MitchellM86} use function models to represent human activities in a simple control system (\eg system shows information about the current state and operator makes relative control actions); Hartson et al.~\cite{Hartson:1990} introduces User Action Notation, a task and user-oriented notation to represent behavioural asynchronous, direct manipulation interface designs; Bolton et al.~\cite{BoltonSB11} establishes an Enhanced Operator Function Model, a generic XML-based notation, to gradually decompose tasks into activities, sub-activities and actions.

Task-analytic models depict human-robot co-existence and highlight the active role of humans in the execution of the task. However, they do not always reduce the overall size of the state spaces of the model~\cite{Bolton2010}, especially when multiple human operators are involved in the execution. Moreover, they do not offer re-usability (i.e., dependent on the case-study scenario and not generalizable) and generation of erroneous human behaviour. 
\section{Probabilistic Human Modelling}
\label{sec:prob}
Another approach to reproduce human non-determinism and uncertainty more vividly are probabilistic models. Unlike deterministic models that produce a single output, stochastic/probabilistic models produce a probability distribution. In theory, a probabilistic human model could be very beneficial for model-based safety assessment in terms of a trade-off between cost and safety.
For example, given that human manifests activity A with probability $P << 0.001\%$, and that A causes hazard H which is very expensive to mitigate, system engineers can save some money by not installing expensive mitigation for H that might practically never happen and settle for a more cost-effective mitigation.
But we must analyse the challenges of probabilistic models too, so let us discuss a few examples.

Tang et al.~\cite{DBLP:journals/corr/TangJHG16} propose a bayesian probabilistic human motion model and argue that human mobility behaviour is uncertain but not random, and depends on internal (e.g., individual preferences) and external (e.g., environment) factors. They introduced an algorithm to learn human motion patterns from collected data about human daily activities from GPS and mobile phone data, and extract a probabilistic relation between current human place and past places in an indoor environment. The first issue about this work, and probabilistic models in general, is collecting a big enough dataset for extracting the correct parameters for a probabilistic distribution. The second issue is that the dataset gathered by GPS and cameras is very coarse-grained for modelling delicate HRC situations; they might capture human moving from one corner of a small workspace to another, but do not capture a situation such as human handing an object to the robot gripper or changing the tool-kit installed on the robot arm.

Hawkins et al.~\cite{7030020} present an HRC-oriented approach, based on a probabilistic model of humans and the environment, where the robot infers the current state of the task and performs the appropriate action. This model is very interesting but is more suited for simulation experiments, thus needs adaptive changes to be compatible with formal verification tools. The model also assumes that human performs everything correctly which is not quite realistic.

Tenorth et al.~\cite{DBLP:conf/icra/TenorthTB13} uses Bayesian Logic Networks~\cite{Jain2011BayesianLN} to create a human model and evaluated their approach with TUM kitchen~\cite{tum} and the CMU MMAC~\cite{Torre2009CMU-MMAC} data sets, both focusing only on full-body movements that exclude many HRC-related activities (e.g., assembling, screw-driving, pick and place, etc).

The Cognitive Reliability and Error Analysis Method (CREAM)~\cite{hollnagel1998cognitive} is a probabilistic cognitive model that is used for human reliability analysis~\cite{DoughertyFragola1988}, therefore, focuses on correct and incorrect behaviour both; it defines a set of modes to replicate different types of human behaviour which have different likelihoods to carry out certain errors. Similarly, Systematic Human Error Reduction and Prediction Analysis (SHERPA)~\cite{embrey1986sherpa} is a qualitative probabilistic human error analysis with a task-analytic direction which contains several failure modes: action errors, control errors, recovery errors, communication errors, choice errors. De Felice et al.~\cite{DEFELICE20161673} propose a combination of CREAM and SHERPA that uses empirical data to assign probabilities to each mode; therefore, the results are strongly dependent on the case study domain and the reliability and amount of data.

In recent years, machine-learning algorithms have been used quite extensively to create probabilistic human models~\cite{DBLP:journals/sensors/ManniniS10,10.1007/978-1-4471-0765-1-28,kim2015interactive}, however having a reliable and large-enough dataset to learn probabilities from remains an issue. The communities (i.e., providers, users) shall develop such datasets for different domains by storing log histories and integrating them into a unique dataset (e.g., a dataset for industrial assembly tasks, a dataset for service robot tasks, etc.). The larger these datasets become, the more reliable they get, the better the human models extracted from them will be.

\section{Human Erroneous Behavior}
\label{sec:err}
Human operators are prone to errors---an activity that does not achieve its goal~\cite{reason1990human}, and one significant source of hazards in HRC applications is human errors. Reason~\cite{reason2000human} metaphorically states that the weaknesses of safety procedures in a system allow for the occurrence of human errors.
Hence, a realistic human model for safety analysis shall be able to replicate errors, too. 
\\
The previous sections explained different possibilities to model human behaviour. They all could be used to model human errors, too, but are not always used as such; some of the papers discussed above, consider errors and others entirely skip them. This challenge is discussed in a separate section, but all of the papers discussed below could be listed in at least one of the previous three sections.

One must first recognize errors and settle for a precise definition for them to model them. Although there is no widely used classification of human errors, some notable references follow.

\begin{figure}
	\centering
	\includegraphics[scale=1]{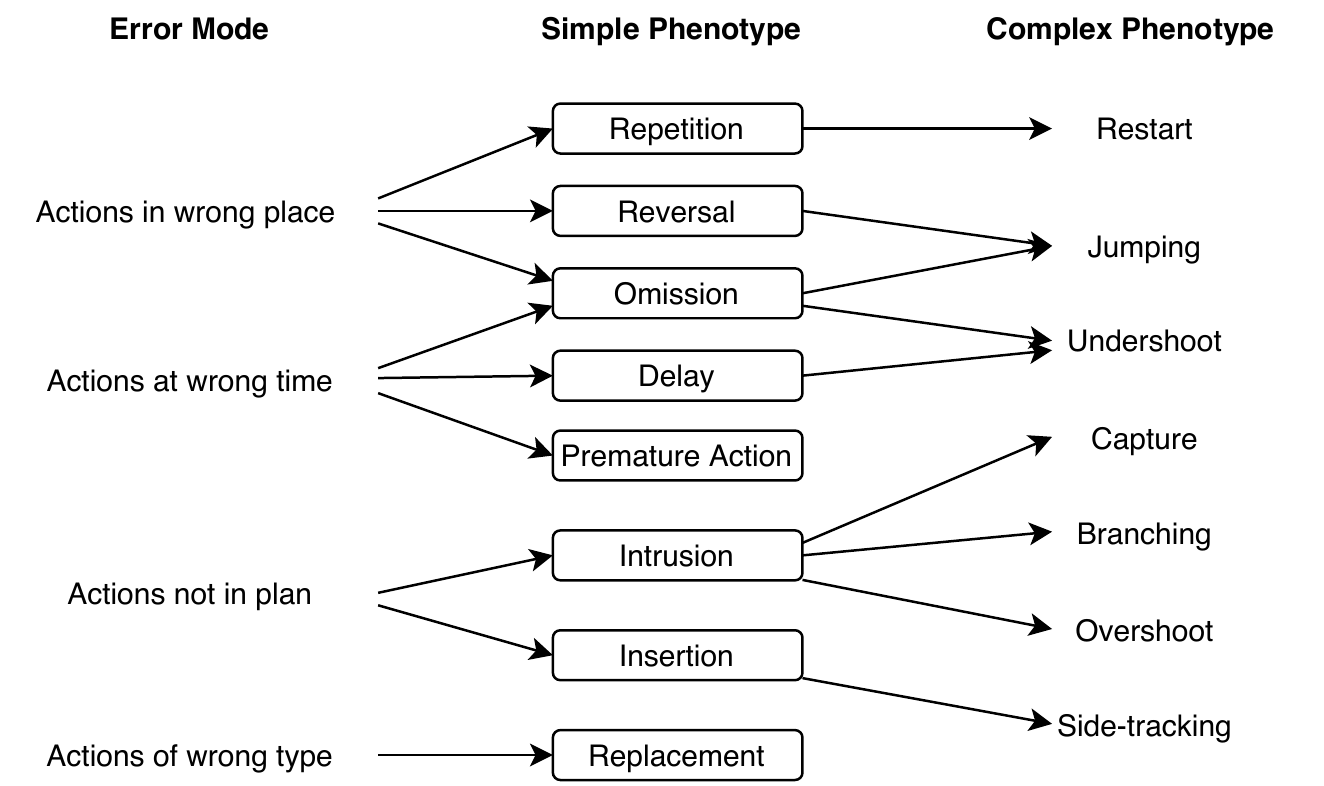}
	\caption{A taxonomy of erroneous actions phenotypes, taken from \cite{Fields}. Wrong place refers to the action's temporal position in the execution sequence, not to a location in the layout; undershooting, which occurs when the action stops too early; sidetracking occurs when a segment of unrelated action is carried out, then the correct sequence is resumed; capturing occurs where an unrelated action sequence is carried out instead of the expected one; branching is where the wrong sequence of actions is chosen; overshooting happens where the action carries on past its correct endpoint by not recognizing its post-conditions.}
	\label{tab:tax-chart}
\end{figure}

Reason~\cite{reason1990human} classifies errors as behavioral (i.e., task-related factors), contextual (i.e., environmental factors) and conceptual (i.e., human cognition).
Shin et al.~\cite{ShinWR06a} divide errors into two main groups; location errors happen when humans shall find a specific location for the task, and orientation errors happen due to humans' various timing or modality to perform a task. Hollnagel~\cite{hollnagel1998cognitive} classifies human errors in eight simple phenotypes: repeating, reversing the order, omission, late or early execution, replacement, intrusion of actions and inserting an additional action from elsewhere in the task; the paper then combines them to get complex phenotypes, as shown in \Cref{tab:tax-chart}, and manually introduces them in the formal model of the model which produces too many false negatives. This issue might be resolved by introducing a probability distribution on phenotypes. 

Kirwan has conducted an extensive review on human error identification techniques in \cite{KIRWAN1998157} and suggested the following as the most frequent version:
\begin{itemize}
	\item Slips and lapses regarding the quality of performance.
	\item Cognitive, diagnostic and decision-making errors due to misunderstanding the instructions.
	\item Maintenance errors and latent failures during maintenance activities.
	\item Errors of commission when the human does an incorrect or irrelevant activity.
	\item Idiosyncratic errors regarding social variables and human emotional state.
	\item Software programming errors leading to malfunctioning controllers that put the human in danger.
\end{itemize}

Cerone et al.~\cite{CeroneLC05} uses temporal logic to model human errors that are defined as: fail to observe potential conflict, ineffective or no response to observed conflict, fail to detect the criticality of the conflict.

As explained above, the error definition is context-dependent and might vary from one domain to another.  It seems that the taxonomy presented in \cite{Fields} is comprehensive enough for possibilities in HRC scenarios.
Nevertheless, the discussion above was a general description of the violation of Norms; we must explore also papers that model the above definitions.

Many of the works on modelling human errors are inspired by one of the two main steps of human reliability analysis~\cite{DoughertyFragola1988} that are error identification and error probability quantification~\cite{MOSLEH2004241}. For example, Martinie et al.~\cite{DBLP:journals/thms/MartiniePFBFS16} proposes a deterministic task-analytic approach to identify errors, while CREAM, HEART \cite{27540}, THERP \cite{swain1983handbook} and THEA \cite{DBLP:conf/interact/PocockHWJ01} are probabilistic methods for modelling human. However, the models generated by these methods need formalization and serious customization because they strongly depend on their case study. Examples of other works with the same issues follow.

\cite{Pan2016,Bolton15} study the impacts of miscommunication between multiple human operators while interacting with critical systems; \cite{basnyat2005task,paterno2002preventing} explore human deviation from correct instructions using ConcurTaskTrees; \cite{RuksenasBCB09} upgrades the SAL cognitive model with systematic errors taken from empirical data.

On the other hand, several works developed formal human models. For example, Curzon and Blandford~\cite{curzon2004formally} propose a cognitive formal model in higher-order logic that separates user-centred and machine-centred concerns. It uses the HOL proof system~\cite{LauO95}  to prove that following the proper design rules, the machine-centred model does not allow for errors in the user-centred model. They experimented with including strong enough design rules in the model, so to exclude cognitively plausible errors by design~\cite{CurzonB02}. This method clearly ignores realistic scenarios in which human actually performs an error.
In another example, Kim et al.~\cite{KimRJW10} map human non-determinism to a finite state automaton. However, their model is limited to the context of prospective control. Askarpour et al.~\cite{ASKARPOUR2019465,DBLP:conf/safecomp/AskarpourMRV17} formalize simple phenotypes introduced by Fields~\cite{Fields} with temporal logic, and integrate the result in the overall system model which is verified against a physical safety property.
\begin{table}
\caption{A summary of the discussed papers, extracted by a snowballing literature review.
\texttimes{} means that the feature is not included, \checkmark means that the feature is included, ? means that the feature has not been explored by the papers of the row but potentially could be addressed by their proposed approach (regardless of the required effort and the resulting efficiency), ``semi'' means that the model has partially formal semantics but is not in a form to be fed to an automated verification tool. When a reproducible model or approach has been used, the name of the approach is mentioned instead of a \checkmark.
The acronyms used here are explained in \Cref{Tab:acronyms}.
}
\begin{tabularx}{\textwidth}{X|X|X|X|X|X|X}
\toprule
modelling Approach & Cognitive & Task-Analytic & Probabilistic & Modelling Errors & Formalised & HRC Compatible
\\\toprule
\cite{laird2012soar} & SOAR & \texttimes & \texttimes & ? & \texttimes & \texttimes
\\\hline
\cite{anderson1996act} & ACT-R & \texttimes & \texttimes & ? & \texttimes & \texttimes
\\\hline
\cite{GallF14,both2007formal}& ACT-R & \texttimes & \texttimes & ? & \checkmark & \texttimes
\\\hline
\cite{HowesY97}& SOAR & \texttimes & \texttimes & ? & \checkmark & \texttimes
\\\hline
\cite{COGS334,RITTER20011,Stocker2012}& PUM  & \texttimes & \texttimes & ? & \checkmark & ?
\\\hline
\cite{Butterworth2000,WertherS05}& PUM  & \texttimes & \texttimes & ? & \checkmark & \texttimes
\\\hline
\cite{BOLTON2019168}& \texttimes  & EOFM & \texttimes & \checkmark & \texttimes & \checkmark
\\\hline
\cite{LI2016318}& \texttimes  & Graphs & \texttimes & \texttimes & \texttimes & \texttimes
\\\hline
\cite{paterno1997concurtasktrees} &\texttimes&CCT&\texttimes&?&semi&\texttimes\\\hline
\cite{MitchellM86} &\texttimes&FSA&\texttimes&\checkmark&\checkmark&\texttimes\\\hline
\cite{Hartson:1990}&\texttimes&UAN&\texttimes&\texttimes&semi&\texttimes\\\hline
\cite{BoltonSB11}&\texttimes&EOFM&\texttimes&\texttimes&\texttimes&\texttimes\\\hline
\cite{DBLP:journals/corr/TangJHG16}&\texttimes&\texttimes&Bayesian&\checkmark&\texttimes&\texttimes\\\hline
\cite{7030020}&\texttimes&\texttimes& DBN &\texttimes&\texttimes&\checkmark\\\hline
\cite{DBLP:conf/icra/TenorthTB13}&\texttimes&\texttimes&BLN&?&\texttimes&\texttimes\\\hline
\cite{DEFELICE20161673} &\texttimes&\texttimes&CREAM and SHERPA&\texttimes&\texttimes&?\\\hline
\cite{DBLP:journals/sensors/ManniniS10,10.1007/978-1-4471-0765-1-28,kim2015interactive}&\texttimes&\texttimes&ML&?&\texttimes&?\\\hline
\cite{Fields}&\checkmark&\checkmark&\texttimes&\checkmark&\texttimes&?\\\hline
\cite{CeroneLC05}&\texttimes&OCM&\texttimes&\checkmark&\checkmark&?\\\hline
\cite{DBLP:journals/thms/MartiniePFBFS16}&\texttimes&\checkmark&\texttimes&
\checkmark&\checkmark&\texttimes\\\hline
\cite{27540}&\texttimes&\checkmark&CREAM and HEART&\checkmark&\checkmark&?\\\hline
\cite{swain1983handbook}&\texttimes&\checkmark&THERP&\checkmark&\checkmark&?\\\hline
\cite{DBLP:conf/interact/PocockHWJ01}&\texttimes&\checkmark&THEA&\checkmark&\checkmark&?\\\hline
\cite{Pan2016}&\checkmark&EOFMC&\texttimes&\checkmark&LTL&\texttimes\\\hline
\cite{Bolton15}&SAL&EOFM&\texttimes&\checkmark&\texttimes&\texttimes\\\hline
\cite{basnyat2005task}&\texttimes&CCT&\texttimes&\checkmark&semi&?\\\hline
\cite{paterno2002preventing}&\texttimes&CCT&\texttimes&\checkmark&\texttimes&?\\\hline
\cite{RuksenasBCB09}&SAL&\texttimes&\texttimes&\checkmark&\texttimes&\texttimes\\\hline
\cite{curzon2004formally}&PUM&\checkmark&\texttimes&\checkmark&Higher-Order Logic&\texttimes\\\hline
\cite{CurzonB02}&PUM&\checkmark&\texttimes&\checkmark&Temporal Logic&?\\\hline
\cite{KimRJW10}&\cite{DoughertyFragola1988}&\texttimes&\texttimes&\checkmark&FSA&\texttimes\\\hline
\cite{ASKARPOUR2019465}&\texttimes&\checkmark&\texttimes&\checkmark&Temporal Logic&\checkmark\\
\bottomrule 
\end{tabularx}	
\label{Tab:sims}
\end{table}
\begin{table}
\caption{Acronyms Explained.
}
\begin{tabularx}{\textwidth}{l|l}
\toprule
PUM & Programmable User Model \\\hline
OCM & Operator Choice Model \\\hline
EOFM & Enhanced Operator Function Model \\\hline
EOFMC &  Enhanced Operator Function Model with Communications  \\\hline
UAN & User Action Notation  \\\hline
CCT & ConcurTaskTrees \\\hline
FSA & Finite State Automaton\\\hline
CREAM & The Cognitive Reliability and Error Analysis Method\\\hline
SHERPA & Systematic Human Error Reduction and Prediction Analysis \\\hline
LTL & Linear Temporal Logic\\\hline
DBN & Dynamic Bayes Network \\\hline
BLN & Bayesian Logic Networks \\\hline
ML  & Machine Learning \\\hline
THERP & Technique for Human Error-rate Prediction\\\hline
THEA & Technique for Human Error Assessment Early in Design\\\hline
HEART & Human error assessment and reduction technique\\\hline
SHERPA & Systematic Human Error Reduction and Prediction Analysis\\
\bottomrule 
\end{tabularx}	
\label{Tab:acronyms}
\end{table}
\section{Conclusions}
\label{sec:con}
This paper reviews the state-of-the-art on modelling human in HRC applications for safety analysis. The paper explored several papers, excluding those on interface applications, and grouped them into three main groups which are not mutually exclusive: cognitive, task-analytic, and probabilistic models. Then it reviewed state-of-the-art on modelling human errors and mistakes which is absolutely necessary to be considered for safety. Human error models expand over the three main groups but have been discussed separately for more clarity. \Cref{Tab:sims} summarizes the observations from which the following conclusions are drawn:
\begin{itemize}
\item Cognitive models are very detailed and extensive. They often originate from different research areas (e.g., psychology), therefore have been developed with a different mentality from that of formal methods practitioners. So, they must be Formalised; but their formalization requires a lot of effort for abstraction and HRC-oriented customization.
They also require specialist training for modellers to understand the models and be able to modify them. 
 \item Among cognitive models, PUM seems to be the one with more available formal instances. It also could be tailored to different scenarios with much less time, effort and training, compared to ACT-R and SOAR.
 \item Task-analytic models offer little reusability; they are so intertwined with the task definition that a slight change in the task might cause significant changes to the model. They also must be first defined with a hierarchical notation for the easy and clear decomposition of the task and then be translated to an understandable format for a verification tool.
 \item Probabilistic models seem to be an optimal solution; they combine either task-analytic or cognitive models with probability distributions. However, the biggest issue here is to have reliable and large-enough data sets to extract the parameters of probability distributions from. It requires the robotics community to produce huge training datasets from their system history logs (e.g., how frequently human moves in the workspace, the average value of human velocity while performing a specific task, the number of human interruptions during the execution, the number of emergency stops, the number of reported errors in an hour of execution, ...).
 \item None of the three approaches above is enough if they exclude human errors. There are several works on outlining and classifying human errors. The best way to have a unified terminology would be for standards organizations in each domain to introduce a list of the most frequent human errors in that domain. It would be possible only if the datasets mentioned above are available. Modelling all of the possible human errors might not be feasible, but modelling those that occur more often is feasible and considerably improves the quality of the final human model.
 \item Error models might introduce a huge amount of false positive cases (i.e., incorrectly reporting a hazard when it is safe) during safety analysis. Thus, probabilistic error models might be the best combination to resolve it.
 \item In the safety analysis of HRC systems, the observable behaviour of humans and its consequences (i.e., how it impacts the state of the system) are very important. However, the cognitive elements behind it are not really valuable to the analysis. One could use them as a black-box that derives the observable behaviour with a certain probability but the details of what happens inside the box do not really matter. Therefore, the cognitive model seems to contain a huge amount of detail that does not necessarily add value to the safety analysis but makes the model heavy and the verification long.
 \item \Cref{Tab:sims} suggests that a combination of probabilistic and task-analytic approaches that model erroneous human behaviour is the best answer for a formal verification method for safety analysis of HRC systems.
\end{itemize}
\bibliographystyle{eptcs}
\bibliography{bib}
\end{document}